\documentclass[sigconf]{acmart}
\usepackage{fancyhdr}
\usepackage[T1]{fontenc}
\usepackage[scaled=0.85]{beramono}
\usepackage{listings}
\AtBeginDocument{%
  \providecommand\BibTeX{{%
    \normalfont B\kern-0.5em{\scshape i\kern-0.25em b}\kern-0.8em\TeX}}}

\copyrightyear{2021} 
\acmYear{2021} 
\setcopyright{acmlicensed}\acmConference[K-CAP '21]{Proceedings of the 11th Knowledge Capture Conference}{December 2--3, 2021}{Virtual Event, USA}
\acmBooktitle{Proceedings of the 11th Knowledge Capture Conference (K-CAP '21), December 2--3, 2021, Virtual Event, USA}
\acmPrice{15.00}
\acmDOI{10.1145/3460210.3493552}
\acmISBN{978-1-4503-8457-5/21/12}

\begin{document}
\fancyhead{}
\title{Marriage is a Peach and a Chalice: \\ Modelling Cultural Symbolism on the Semantic Web} 
%

\author{Bruno Sartini}
\email{bruno.sartini3@unibo.it}
\orcid{0000-0002-9152-4402}
\affiliation{%
  \institution{University of Bologna}
  \streetaddress{}
  \city{Bologna}
  \state{}
  \country{Italy}
  \postcode{40126}
}

\author{Marieke van Erp}
\email{marieke.van.erp@dh.huc.knaw.nl}
\orcid{0000-0001-9195-8203}
\affiliation{%
  \institution{KNAW Humanities Cluster}
  \streetaddress{Oudezijds Achterburgwal 185}
  \city{Amsterdam}
  \state{}
  \country{The Netherlands}
  \postcode{1012 DK}
 }

\author{Aldo Gangemi}
\email{aldo.gangemi@unibo.it }
\orcid{0000-0001-5568-2684}
\affiliation{%
  \institution{University of Bologna}
  \streetaddress{}
  \city{Bologna}
  \state{}
  \country{Italy}
  \postcode{40126}
 }

\renewcommand{\shortauthors}{Sartini et al.}

\begin{abstract}
In this work, we fill the gap in the Semantic Web in the context of Cultural Symbolism. Building upon earlier work in \cite{sartini_towards_2021}, we introduce the Simulation Ontology, an ontology that models the background knowledge of symbolic meanings, developed by combining the concepts taken from the authoritative theory of Simulacra and Simulations of Jean Baudrillard with symbolic structures and content taken from ``Symbolism: a Comprehensive Dictionary'' by Steven Olderr. We re-engineered the symbolic knowledge already present in heterogeneous resources by converting it into our ontology schema to create HyperReal, the first knowledge graph completely dedicated to cultural symbolism. A first experiment run on the knowledge graph is presented to show the potential of quantitative research on symbolism.
\end{abstract}





\maketitle

\pagestyle{fancy}
\section{Introduction}
Symbols are strongly related to human expression and communication, for this reason, they can be found in many contexts such as art, literature, music, as well as more recently in movies and commercials. However, the knowledge about symbolism, which encompasses interpretations of cultural objects, along with the information about canonical symbols from different cultures is mostly stored in unstructured sources such as encyclopedias and dictionaries of symbols. Accessing and processing this information quantitatively is therefore currently not possible. This paper presents a novel approach in modeling cultural symbolic knowledge. We focus on conventional symbolic knowledge as expressed by experts or in a dictionary as opposed to symbolic knowledge expressed in a specific work of art. 

Symbols are complex objects: the same symbol might convey different meanings depending on the cultural or artistic context in which it is found \cite{frazer_golden_1922,roberts_encyclopedia_2013}. Furthermore, different cultures might convey the same meaning with different symbols. This is for example the case for the concept of marriage, that is expressed in the Chinese Culture by the symbol of a peach \cite{otto_mythological_2014}, but in the Celtic context, the concept of marriage is expressed by the symbol of a Chalice \cite{olderr_symbolism_2012}. 

Linked open data sets in the cultural heritage domain currently lack domain-specific ontologies that are able to express symbolic relationships in a structured way. \cite{sartini_towards_2021} explains how the lack of semantic models on the domain of symbolism negatively affects the interlinking of resources through their symbols and shows promising results in the application of a prototype ontology schema to a small set of data. Given the impact of symbols, not only on the representation, comparison and evolution of human cultures, but also in common expressions, we are bringing symbolic knowledge, that has been accumulated throughout centuries, into a machine readable and open format to foster quantitative research on this topic.

Specifically, our goal is to fill the gaps on cultural symbolism that currently exist in linked open data through: (i) the Simulation Ontology, a newly developed ontology that conceptualises symbolic relationships; and (ii) HyperReal, a knowledge graph that contains data from heterogeneous sources converted into the Simulation Ontology schema. The Simulation Ontology and HyperReal are interoperable with current knowledge graphs, open resources and utilisable to encode symbolic knowledge contained in written unstructured sources. 


\section{State of the art}
\label{section2}
In this section, we analyse the coverage of existing ontologies and knowledge graphs with respect to their symbolic content and whether these are modelled as instances of symbolism, and relationships that link a symbol to its symbolic meaning. 

General domain knowledge bases, such as Wikidata \cite{wikidatapaper} and DBpedia \cite{dbpediapaper} contain some properties that link a resource to its symbolic meaning. In Wikidata, the property \textsc{P4878}\footnote{\url{https://www.wikidata.org/wiki/Property:P4878}} (\textit{symbolizes}) is a qualifier for statements in which the property P180\footnote{\url{https://www.wikidata.org/wiki/Property:P180}} (\textit{depicts}) is used and it should represent the symbolic meaning of elements depicted in a work of art. Out of more than half a million elements that have been linked to a work of art using the P180 property, only 313 have a \textit{symbolizes} qualifier.\footnote{This data was extracted through two queries in the Wikidata SPARQL portal: the query \url{https://w.wiki/3z56} counts the number of times P4878 is used as a qualifier, the query \url{https://w.wiki/3z5A} counts the number of elements linked with the property P180.}

DBpedia uses the property \texttt{dbp:symbol}\footnote{\url{http://dbpedia.org/property/symbol}} to link a concept to a symbol that represents it. The range of the \texttt{dbp:symbol} property is very general, as the property is used both to describe cultural symbols, for instance thunderbolts as the symbol of Zeus\footnote{\url{https://dbpedia.org/page/Zeus}} but also to express that ``metro'' is the symbol of the Atocha railway station in Madrid. \footnote{\url{https://dbpedia.org/page/Madrid_Atocha_railway_station}}

Iconclass~\cite{Couprie1983IconclassAI} is a classification system used for attributing of subjects to works of art. Each subjects is given a specific code according to the hierachical structure of the system. Specific codes are also given for attributing subjects that appear with symbols, for example, the Iconclass code 11HH(MARY MAGDALENE)\footnote{\url{http://iconclass.org/rkd/11HH\%28MARY\%20MAGDALENE\%290/}} is defined as ``the penitent harlot Mary Magdalene; possible attributes: book (or scroll), crown, crown of thorns, crucifix, jar of ointment, mirror[...]''. In this sense, Iconclass contains relevant symbolic information for western art subjects, but its linked open data version consists of a hierarchical SKOS vocabulary. Our work, compared to Iconclass, uses linked open data semantic web technologies to express more specific relationships between symbols and their meaning. 

CIDOC CRM\cite{cidoc} is a popular conceptual model for describing museum and cultural heritage artefacts. Its event-centric structure makes it possible to attribute a particular meaning to a specific cultural object. This possibility was extended by the VIR ontology \cite{carboni_ontological_2019}, based on CIDOC, using a specific property \textit{symbolize}\footnote{\url{https://ncarboni.github.io/vir/\#K14_symbolize} url not working for every browser. Persistent URI of vir ontology: \url{http://w3id.org/vir}
} to link certain elements in a representation of an artwork to their symbolic meaning. The model was enriched to describe complex iconological case studies of interpretation \cite{baroncini2021modelling}. However, the dependency of the interpretation acts makes symbolic relationships encoded using this ontology solely based on the single work of art (i.e. not generalized). In other words, currently CIDOC and VIR permit to describe symbols that are only valid in the context of the work of art in which they are found as a result of an interpretation.  The purpose of our work is to first introduce a model that can express general background symbolic knowledge without covering the hermeneutical act of interpretation.
Parts of CIDOC model will be re-used to develop an additional ontology of iconographic interpretations in future work.


Gartner \cite{gartner_towards_2020} proposes an ontology to conceptualize iconographical recognition of subjects in artworks according to Panofsky's second level of interpretation.\footnote{Panofsky's second level of iconographic interpretation is about the subject matter of works of art. In this level characters, symbols, places, events, allegories and stories are associated to the artistic motifs present in the work of art \cite{panofsky_studies_1962}.} Among the classes that are conceptualized in Gartner's ontology there is a mention of symbols as recognizing elements for specific work of art subject. Unfortunately, the ontology has not been released yet.

Symbolic information can also be found in Wordnet \cite{fellbaum_Wordnet_1998}, a lexical database of words and relationships between them that has been under development by cognitive linguists since 1985. The dollar sign\footnote{ \url{http://wordnet-rdf.princeton.edu/id/06834465-n}} for example is defined as ``a symbol of commercialism or greed'', but it does not contain structured information to distinguish between its meaning as denoting a currency and as a symbol of greed. Only in some cases, a meaning is described as figurative, although only in its definition. For instance, for Albatross\footnote{\url{http://wordnet-rdf.princeton.edu/id/05697450-n}} the definition ``(figurative) something that hinders or handicaps'' appears in the textual description of the entry. Wordnet's aim is not to distinguish between symbolic and literal meanings, but the information that can be found in this resource can be extracted and re-engineered in order to highlight the former from the latter. As explained in Section \ref{kbcreation}, Wordnet's symbolic data was ingested in our knowledge graph.
 

\section{Ontology design and implementation}
\label{designsec}

In this section, we explain the design stages of the ontology together with the first implementation of the ontological model through a conversion of data from multiple heterogeneous sources.

The ontology was designed following the agile ontology development methodologies of SAMOD\cite{samod} and extremeDesign\cite{xd}. The design of the ontology was completed in three SAMOD iterations. The first iteration covers the aspects related to the top-down modelling (subsection \ref{section3.1}), the second and third iterations cover aspects related to the data-driven modelling (subsection \ref{subsection3.3}). For each iteration, the following documentation items were generated: (i) a motivating scenario; (ii) an example of data that covers the scenario in natural language; (iii) a series of competency questions in natural language with the expected results related to the example; (iv) a glossary that contains descriptions of the terms used in the motivating scenario; (v) an ABOX that contains the example encoded using the ontology schema as a  turtle\footnote{\url{https://www.w3.org/TR/turtle/}} serialisation, (vi) unit tests that test the competency questions.
The full documentation is available at \url{https://www.w3id.org/simulation/development}.

\subsection{Top-down Ontology Development}
\label{section3.1}
Eco \cite{eco_semiotica_1996} presents an overview of different theories on symbols and symbolism. Compared to the highly debated topic (according to~\cite{eco_semiotica_1996}) of what symbols are and what can be considered a symbol, there is less discussion about the relationship that links the symbol to what it symbolizes. This relationship is partly related to Eco’s view on \textit{ratio difficilis}: the meaning of the expression is connected with its content and the ``deciphering'' or inference of this meaning depends on whether the correlation between the expression and content is based on pre-existing rules or it is something personally created by the issuer of the symbolic linking. 

Of the theories in Eco's overview, Baudrillard’s Simulation and Simulacra \cite{baudrillard_simulacra_1981}, mentions the terms simulation, simulacrum and reality counterpart as the three main elements of a symbolic relationship. 
A \textit{simulation} is intended as the relationship between a symbolic element and its meaning. The meaning expressed in a simulation is different from the literal meaning of the element. Lion as a symbol of courage \cite{olderr_symbolism_2012} is a simulation, lion as ``a large wild animal of the cat family with yellowish-brown fur that lives in Africa and southern Asia'' \cite{cambridgeuniversitypress} would not be considered as a simulation.
The \textit{simulacrum} is the symbolic element, it is the representation of something else. The \textit{reality counterpart} is the ``something else''  represented by the \textit{simulacrum},  not the literal meaning of the simulacrum itself.  An olive branch (simulacrum) represents ``peace'' (reality counterpart). The simulation ``olive branch-peace'' is the symbolic relationship that links these two elements. Simulations are not universally valid, some only exist in specific settings or contexts. An owl is the symbol of death in Hindu, Japanese, and Mayan contexts. That means that the simulation \textit{owl-death} exists in those contexts. On the other hand, in a Siberian context, owls are symbols of helpful spirits~\cite{olderr_symbolism_2012}.

\begin{figure}[t]
\includegraphics[width=\columnwidth]{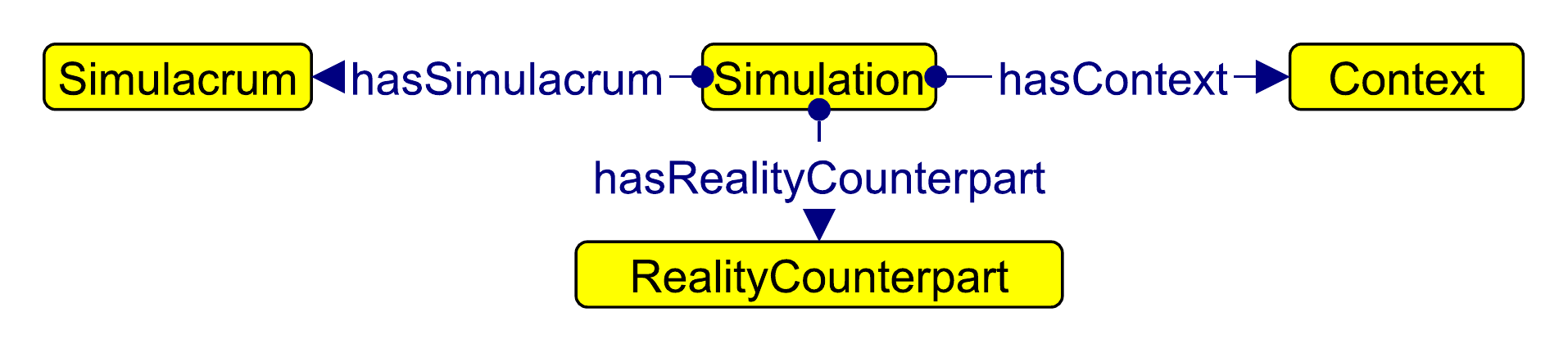}
\caption{Simulation Pattern - Classes and Properties}
\label{fig:sim1}
\end{figure}


From the analysis of Simulation theory, the following competency questions emerged: (Q1.1) What are the reality counterparts of the simulations that have a specific simulacrum? (Q1.2) What are the simulations that exist within a certain context? (Q1.3) What are the simulations in which a certain element participates as either the simulacrum or the reality counterpart? (Q1.4) What are the simulacra that share the same reality counterpart in their respective simulations and what is the context in which their simulations exist?
A negative competency question\footnote{Intended as a competency question which should not return any result because it asks for something which does not follow the logical structure of the ontology.} was also formulated: (Q1.5) Are there simulations that have multiple simulacra?

As the ontology was not intended as an in-depth description of the philosophical theories of Baudrillard, only the main concepts of his work were reused for the conceptualization of symbolic meanings. To establish a solid foundation in existing good practices, \cite{10.5555/3099984} three off-the-shelf ontology patterns were reused to design the OWL2 version of the Simulation ontology: \textit{situation}\footnote{\url{http://www.ontologydesignpatterns.org/cp/owl/situation.owl}} provides a general structure and vocabulary for N-ary relations; \textit{semiotics}\footnote{\url{http://www.ontologydesignpatterns.org/cp/owl/semiotics.owl}} founds Baudrillard's simulations as semiotic acts. An expression (here a simulacrum) denotes a reference (here a reality counterpart), with an interpreted meaning (here muted, but actualizable in the context of a specific interpretation act e.g. in iconology); and \textit{information realization},\footnote{\url{http://www.ontologydesignpatterns.org/cp/owl/informationrealization.owl}} which lets us distinguish information objects from their realization or manifestation (here relevant for distinguishing conventional simulations from simulations interpreted e.g. for a specific work of art). 
In other words, the simulation ontology for conventional symbolic meaning presented here holds between pure information objects (e.g. a lion prototype), and concepts (e.g. force), or stereotyped individuals (e.g., a generic Persia).\footnote{This is contrasted by using different ontologies that implement simulation occurrences, e.g. a specific interpretation of Persepolis' Achaemenid Persian relief with the Sign of Lion, as a realized simulacrum of Achaemenid Persia power.} The design pattern for simulations is presented in Figure \ref{fig:sim1}.

\begin{figure}[t]
\includegraphics[width=\columnwidth]{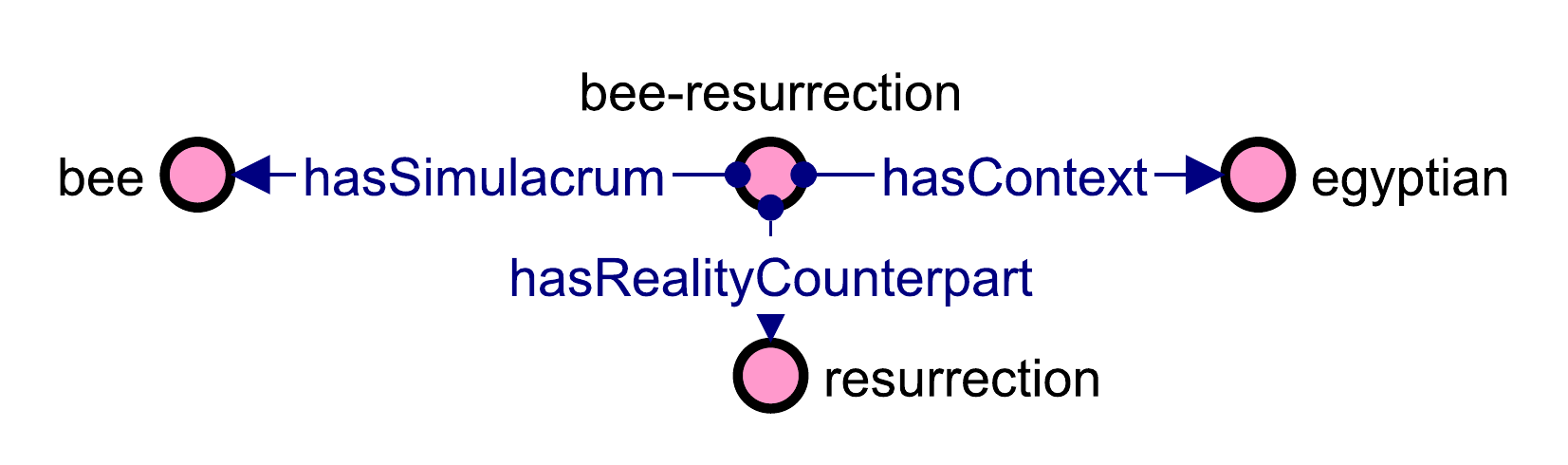}
\caption{Simulation Pattern - Bee-resurrection simulation example}
\label{fig:sim2}
\end{figure}

Simulacra, reality counterparts, and contexts of symbolic meanings are linked in this pattern through the n-ary relationship (situation) class Simulation. For example, in an Egyptian context, bees signify resurrection. This can be modelled through the \textit{hasSimulacrum}, \textit{hasRealityCounterpart} and \textit{hasContext} properties in our model, as shown in Figure \ref{fig:sim2}.

\begin{figure*}[ht]
\includegraphics[width=12cm]{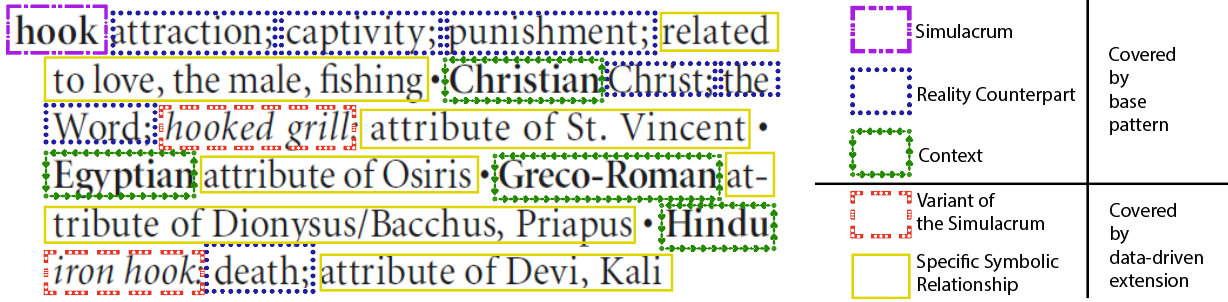}
\caption{Fitting the simulation pattern over Olderr's dictionary "hook" entry}
\label{fig:sim4}
\end{figure*}

To evaluate the initial version of the developed pattern, we qualitatively compare its ontological structure to the structure of a dictionary of symbols. We selected Olderr's~\cite{olderr_symbolism_2012} dictionary of symbols for this task because it offers more than 40,000 symbolic meanings. Moreover, its structure is similar to a dictionary of synonyms and antonyms: for each symbol, a list of potential symbolic meanings is provided along with the context in which those symbolic meanings are valid, often without any additional information. As we will discuss in the conversion Section \ref{kbcreation} the repetitive patterns in the structure of this dictionary facilitate the automatic conversion of its data into the ontology schema. An example of fitting the pattern to the structure of the dictionary is shown in Figure \ref{fig:sim4}. The example shows the ``hook'' entry of the dictionary. Hook translates in our ontology as the simulacrum. The terms in green (e.g. Christian) represent the contexts in which some hook simulations are valid. The terms in blue represent reality counterparts of simulations that have hook as a simulacrum. However, some reality counterparts (in yellow) are introduced by phrases such as ``related to'', ``attribute of'' that suggest a specific type of simulation. Also, the terms in red represent some variants of the ``hook'' entry that have their own symbolic meaning. Both variants and specific simulations were addressed in the design extension of the pattern. The conceptualized classes and properties derived from this qualitative comparison are described in the Subsection \ref{subsection3.3}.

\subsection{Data-driven Ontology Enrichment}
\label{subsection3.3}
To cover concepts not dealt with in Baudrillard's Simulation theory, we add 7 new properties and 9 new classes via a bottom-up approach. 

Simulations are linked to the sources that support their existence, opening up new competency questions: (Q2.1) What are the simulations and respective reality counterparts that have the same simulacrum but a different source? (Q2.2) What are the contexts of the simulations listed in a specific source? (Q2.3) What are the sources of a specific simulation? Another negative competency question was formulated: (Q2.4) Are there simulations that do not have a source?
The source of the simulation was included in the ontology using the class \textit{source}. PROV-O\cite{lebo2013prov} is the W3C standard ontology to express provenance. The property  of PROV-O \textit{wasDerivedFrom}\footnote{\url{https://www.w3.org/TR/prov-o/\#wasDerivedFrom}} is used to link a simulation to its source.

In Olderr's dictionary of symbols \cite{olderr_symbolism_2012}, simulacra and reality counterparts can have variants that can belong to different simulations compared to the original simulacra or reality counterparts. Variants in this ontology are either represented as (i) a narrower concept than the original simulacrum or reality counterpart, such as \textit{night bird}, which in Olderr's dictionary is a variant of \textit{bird}, (ii) as a set of things of which the original simulacrum is a part of, such as in the case of \textit{black and white}, which is a variant of \textit{black} in the same dictionary, or (iii) as the simulacrum put in a specific situation such as \textit{bloodstone placed in a glass of water during a drought} as a variant of \textit{bloodstone}.

Variants are conceptualized in the ontology with the introduction of the property \textit{hasVariant} that links either a simulacrum or reality counterpart to its variant (which can be either another simulacrum or a reality counterpart). We test this property using  additional competency questions to handle the conceptualization of variants: (Q3.1) What are the variants of a certain element? (Q3.2) What are the reality counterparts of the simulations with a specific simulacrum or its variants?

In Olderr's dictionary, Simulations can be specialized according to the specific symbolic relationship that links a simulacrum with its reality counterpart. If a simulacrum is the emblem of a certain reality counterpart, or an allusion, this specific relationship can be expressed through specialized simulations. Olderr \cite{olderr_reverse_2005} provides some definitions for specific symbolic relationships such as the Allusion, described as \begin{quote}
    [...]a reference to an historical person or event, or an artistic or literary work. To have an ``albatross around one's neck'' is an allusion to Coleridge’s poem ``The Rhyme of the ancient Mariner''.
\end{quote}
or the Association, described as
\begin{quote}
    [...]something linked in memory or imagination, or by correlation or an analogy with an object, idea, person, or event. The letter ``A'' is associated with beginning.
\end{quote}
Although we suggest potential users of the ontology to follow his definitions, given the highly subjective topic of the ontology, we tried to not limit the use of the classes through logical constrains, as seen in section \ref{sec:ontologyax}.  In some specialized simulations, the reality counterpart might not be the exact symbolic meaning of a simulacrum. In these cases, the reality counterpart might be something that is prevented, elicited, restored by the simulacrum in a symbolic way. Finally, some simulations might have a reality counterpart that represents the symbolic meaning of the simulacrum, and an additional one that represents something that is prevented, elicited, restored by the simulacrum. For instance, in an Arabian context, an agate is seen as a charm for healthy blood~\cite{olderr_symbolism_2012}. Therefore, the resulting simulation is agate-charm-healthyBlood where agate is the simulacrum, charm is a reality counterpart, healthy blood is an elicited reality counterpart and Arabian is the context. Given that a reality counterpart, such as \textit{healthy blood}, does not change its identity whether it is prevented, elicited, or generally symbolically meant by a Simulacrum, we decided to introduce specific reality counterpart relationships as sub-properties of \textit{hasRealityCounterpart} and not as sub-classes of \textit{RealityCounterpart}.

To address specific simulations, we formulate the following competency questions: (Q3.3) What are the simulations in which the simulacra are seen as symbolical protection against reality counterparts? (Q3.4) What are the simulations that have a specific reality counterpart and other additional reality counterparts, and what specific relationship link those simulations to their reality counterparts? (Q3.5) What are the simulations and their respective simulacra, contexts and reality counterparts in which their simulacrum is a symbolical cure for their reality counterpart?

The newly added classes and properties are summarized in Table \ref{table:specificsimulations}. Examples of use, and a glossary containing definitions of all classes and properties can be found in the ontology development documentation.

\begin{table}[ht]
\begin{tabular}{|l|p{4cm}|}
\hline
\textbf{{Class}} & \textbf{{Specific reality counterpart property}} \\ \hline
Association Simulation                   & no specific property                                                     \\ \hline
Correspondence Simulation                & no specific property                                                     \\ \hline
Manifestation Simulation                 & no specific property                                                     \\ \hline
Relatedness Simulation                   & no specific property                                                     \\ \hline
Attribute Simulation                     & no specific property                                                     \\ \hline
Allusion Simulation                      & no specific property                                                     \\ \hline
Protection Simulation                    & preventedRealityCounterpart                                              \\ \hline
Emblematic Simulation                    & no specific property                                                     \\ \hline
Healing Simulation                       & healedRealityCounterpart                                                 \\ \hline
no specific simulation class             & restoredRealityCounterpart                                               \\ \hline
no specific simulation class             & easedRealityCounterpart                                                  \\ \hline
no specific simulation class             & elicitedRealityCounterpart                                               \\ \hline
\end{tabular}
\caption{Specific Simulation and related reality counterpart properties}
\label{table:specificsimulations}
\vspace{-.6cm}
\end{table}

\subsection{Ontology Axiomatisation}
\label{sec:ontologyax}

The simulation ontology uses OWL logical axioms to specify its conceptualisation, to enable logical inferences on the symbolic knowledge graph, and to perform automated classification and consistency checking. 

Besides the basic classes and properties that have been introduced in the previous sections, and their subsumption and domain/range axioms, we exemplify some OWL axioms that perform restrictions on the main classes, written in Manchester Syntax\cite{manchester}. 
\begin{table*}[ht]
\begin{tabular}{|l|r|r|r|r|r|}
\hline
\textbf{Source}                & \textbf{\# of Simulacra} & \textbf{\# of Rc} & \textbf{\# of Contexts} & \textbf{\# of Simulations} & \textbf{\# of Triples} \\ \hline
Olderr's dictionary of symbols & 8,900                     & 17,959             & 303                     & 37,647                      & 450,679                 \\ \hline
DBpedia                        & 3,024                     & 782              & 36                      & 3,727                      & 47,696                  \\ \hline
Wordnet                        & 61                       & 76                & 5                       & 81                         & 1,191                   \\ \hline
\textbf{Total}                 & \textbf{11,663}           & \textbf{18,676}    & \textbf{323}            & \textbf{41,416}             & \textbf{498,525}        \\ \hline
\end{tabular}
\caption{Knowledge Graph statistics}
\label{table:generalnumbers}
\end{table*} 

\begin{itemize}
    \item Simulation: \begin{verbatim} 
    hasContext some Context
    hasRealityCounterpart some RealityCounterpart
    hasSimulacrum exactly 1 Simulacrum
    wasDerivedFrom some Source
\end{verbatim}
\item Healing Simulation: \begin{verbatim}
    healedRealityCounterpart exactly 1
    RealityCounterpart
\end{verbatim}
\item Protection Simulation: \begin{verbatim}
    preventedRealityCounterpart exactly 1
    RealityCounterpart
\end{verbatim}
\end{itemize}

We also introduce a property chain to provide a direct relation facility between simulacra and reality counterparts:

\begin{verbatim} 
    isSimulacrumOf o hasRealityCounterpart 
    SubPropertyOf symbolicMeaning
\end{verbatim}

Further details can be found in the ontology documentation.\footnote{\url{https://www.w3id.org/simulation/docs}}

\subsection{Creating the Symbolic Knowledge Graph}
\label{kbcreation}
The knowledge base was developed by extracting and converting data from different sources: DBpedia, Wordnet and Olderr’s \textit{Symbolism: a comprehensive dictionary}. 

\vspace{-.2cm}
\subsubsection*{DBpedia Conversion}
We use the DBpedia SPARQL endpoint\footnote{\url{https://dbpedia.org/sparql}} to retrieve resources related to the property \textit{dbp:symbol} and for resources of type \textit{skos:Concept}\footnote{\url{http://www.w3.org/2004/02/skos/core\#Concept}} that contain the string "symbol" in their label and that have these concepts as their subject (using the property \textit{dct:subject}.\footnote{\url{http://purl.org/dc/terms/subject}})
For instance, the resource \textit{dbr:Eagle}\footnote{\url{http://dbpedia.org/resource/Eagle}} is the subject of \href{http://dbpedia.org/resource/Category:National_symbols_of_Liechtenstein}{dbc:National\_symbols\_of\_Liechtenstein}.\footnote{\url{http://dbpedia.org/resource/Category:National_symbols_of_Liechtenstein}}

To filter out station signs and chemical codes that are also modelled using the \textit{dbp:symbol} property, but are purely iconic tools, we exclude resources of type Railway Stations and Public Companies. On the one hand, in the triple \textit{?subject dbp:symbol ?object}, ?subject becomes our reality counterpart, and ?object the simulacrum. The \textit{?type
} of the same subject, in the triple \textit{?subject rdf:type ?type} was considered as the context of the simulation. 

On the other hand, in the triple \textit{?subject dct:subject ?object}, \textit{?subject} became the simulacrum and \textit{?object} (cleaned of the initial parts such as ``National Symbol of'' or ``Symbol of'') became the reality counterpart. The contexts,
in this second case were not extracted for these triples, therefore \textit{General or Unknown} was set as the context for every simulation. For instance, in the triple \textit{dbr:Eagle dct:subject dbc:dbc:National\_symbols\_of\_Liechtenstein }, \textit{Eagle} became the simulacrum and \textit{Liechtenstein} became the reality counterpart in a Simulation with a General or Unknown context. Our DBpedia conversion yielded 3,727 simulations expressed in 47,696 triples. 

\vspace{-.2cm}
\subsubsection*{WordNet conversion}
From Wordnet, we extract Symbol and emblem synsets\footnote{\url{http://Wordnet-rdf.princeton.edu/id/05773412-n} and \url{http://Wordnet-rdf.princeton.edu/id/06893714-n} respectively} and their hyponyms, as well as synsets that contain phrases such as \textit{symbol of} or \textit{emblem of} in their definition. For each element, we extract its label and definition. The natural language definition of each synset was processed to extract potential contexts and symbolic meanings. For instance, the Penelope synset\footnote{\url{http://wordnet-rdf.princeton.edu/id/09616318-n}} has "Penelope" as a label and "(Greek mythology) the wife of Odysseus and a symbol of devotion and fidelity[...]" as the definition. The label of the synset was converted into the simulacrum and we extracted the context and reality counterparts from the definition. In this case, two simulations were created from this synset, \textit{penelope-fidelity} and \textit{penelope-devotion}, both with \textit{greekMythology} as a context. The \textit{penelope} simulacrum was then linked to its original WordNet synset with the property \texttt{owl:sameAs}.
Our WordNet conversion process yielded 81 simulations captured in 1,191 triples.
\vspace{-.2cm}
\subsubsection*{Olderr's Dictionary Conversion}
We used the markup such as boldface in Olderr's dictionary to distinguish between different lemmas, contexts and variants and converted this to RDF. A simulation URI was formed by joining the simulacrum and reality counterpart labels with a hyphen. The style consistency in the original document made the conversion process almost completely automatic. Some manual corrections had to be done after the automatic conversion in the rare cases of mistakes in the source.\footnote{\url{https://github.com/br0ast/simulationontology} lists examples of manual corrections} A list of bigrams such as \textit{related to} and \textit{protection from} were used to create \textit{Relatedness} and \textit{Protection} simulations, respectively. 37,647 simulations were generated from the dictionary captured in 450,679 triples.
~\\

For every converted simulation, the respective sources (either DBpedia, Wordnet or Olderr's dictionary) have been added using the \texttt{prov:wasDerivedFrom} property. 

\begin{figure*}[ht]
\includegraphics[width=.7\textwidth]{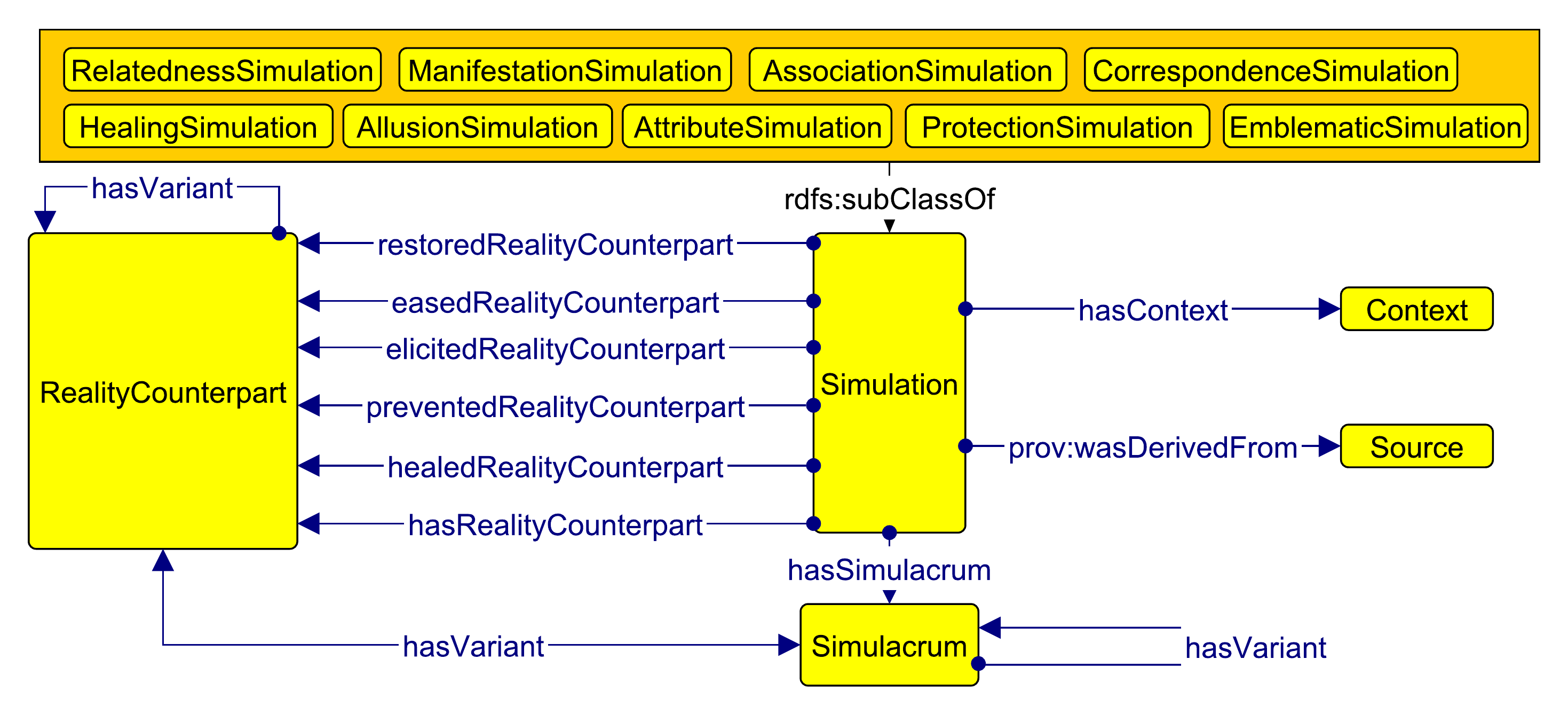}
\caption{Simulation Ontology - Classes and Properties}
\label{fig:sim3}
\end{figure*}

\subsubsection*{Quality Control} After the conversion, we performed several error analysis cycles to assess the knowledge graph's quality, we adjusted the algorithms, and reconverted the knowledge graph where needed. The most common errors were instantiations of some simulacra or reality counterparts that were not linked to any simulation. Other errors occurred during automatic URI generation or in the labeling of elements. Specifically for Olderr's dictionary, some errors were due to editing errors in the original source. In those cases, the txt version of the dictionary was manually corrected.
~\\

Table \ref{table:generalnumbers} summarizes the number of resources that were generated by the conversion.\footnote{The totals are not the exact sum of the previous numbers because some simulacra, reality counterparts, contexts and simulations are shared by different sources.} An evaluation of the conversion algorithms can be found in Section \ref{sec:conversionevaluation}.

\subsection*{Ontology and knowledge graph release}

The w3id service was used to obtain persistent URIs. The current version of the Simulation Ontology is available at \url{https://www.w3id.org/simulation/docs}. Its structure is shown in Figure \ref{fig:sim3}.

Due to copyright issues, the public version of HyperReal contains only the converted data from DBpedia and WordNet. The knowledge graph is available at \url{https://www.w3id.org/simulation/data/} and includes the import axiom for the ontology schema.

Finally, \url{https:www.w3id.org/simulation/development} and \url{www.w3id.org/simulation/code} are the persistent URIs that lead respectively to the ontology development GitHub repository and the scripts used to create HyperReal and our analyses (see the experiment in section \ref{sec:usecase}) over it.

\section{Evaluation}
\label{sec:evaluationandrelease}

In this section, we discuss the evaluation of the ontology via competency questions and an internal consistency check as well as the evaluation of the automatic conversion of Olderr's dictionary.  

\subsection{Competency Question evaluation}
The ontology requirements were expressed through competency questions mentioned in Section \ref{designsec}. For each iteration of the ontology, we executed unit tests to verify that the results of the SPARQL query that formalizes the competency question matched the expected results of those questions in a toy dataset. All unit tests are available from our Github repository in the form of Jupyter notebooks to promote the reproducibility of our results.\footnote{The jupyter notebooks for testing the CQs are available for each SAMOD iteration, respectively in: \href{https://github.com/br0ast/simulationontology/blob/main/development/1/Competency\%20Questions\%20Test\%201.ipynb}{1}, \href{https://github.com/br0ast/simulationontology/blob/main/development/2/Competency\%20Questions\%20Test\%202.ipynb}{2}, \href{https://github.com/br0ast/simulationontology/blob/main/development/3/Competency\%20Question\%20Test\%203.ipynb}{3}.}

All competency questions matched with their expected results. For instance, in CQ2.2 ``What are the simulations and respective reality counterparts and sources that have the same simulacrum but a different source?'' the expected results for the toy dataset are the simulations (i) \textit{olive-fertility} with \textit{fertility} as a reality counterpart and \textit{dictionaryOfSymbols1} as a source and (ii) \textit{olive-immortality} with \textit{immortality} as a reality counterpart and \textit{dictionaryOfSymbols2} as a source. Listing \ref{lst:sparqlcq} shows the formalized query in SPARQL. The retrieved results from this query match the expected results.

\begin{lstlisting}[captionpos=b, caption=CQ2.2 Formalization in SPARQL, label=lst:sparqlcq,
   basicstyle=\ttfamily,frame=single]
PREFIX ex: <https://example.org/> 
PREFIX sim: <https://w3id.org/simulation/ontology/>
PREFIX prov:<http://www.w3.org/ns/prov#>

SELECT distinct ?simulation ?rc ?source WHERE {
?simulation1 sim:hasSimulacrum ?simulacrum ;
             prov:wasDerivedFrom ?source1 .
?simulation2 sim:hasSimulacrum ?simulacrum ;
             prov:wasDerivedFrom ?source2 .
filter (?simulation1 != ?simulation2 && 
?source1 != ?source2) .
?simulacrum sim:isSimulacrumOf ?simulation .
?simulation sim:hasRealityCounterpart ?rc ;
            prov:wasDerivedFrom ?source .}
\end{lstlisting}

\begin{table*}[t]
\begin{tabular}{|l|l|l|l|}
\hline
\textbf{Element}   & \textbf{Precision} & \textbf{Recall} & \textbf{F$_1$} \\ \hline
Simulation         & 0.96                  & 0.97              & 0.97               \\ \hline
Simulacrum         & 0.97                  & 0.98              & 0.98               \\ \hline
Reality counterpart & 0.96                  & 0.97               & 0.97               \\ \hline
Context            & 0.97                  & 0.98               & 0.98               \\ \hline
Type of Simulation & 0.94                & 0.95               & 0.94                \\ \hline
Variant            & 0.97                  & 0.98               & 0.98               \\ \hline
Average            & 0.96                   & 0.97                & 0.97               \\ \hline
\end{tabular}
\caption{Conversion algorithms evaluation metrics (reporting micro averages)}
\label{table:tablemetrics}
\end{table*}

\subsection{Automatic ontology evaluation}


Foops! \cite{foops} is a web application that evaluates ontologies by verifying that they comply with FAIR principles~\cite{fairprinciples}. According to Foops!, our ontology scores 77\%. The tool highlighted the ontology has not been inserted in a public ontology metadata registry of ontology such as LOV~\cite{lov} and that it currently lacks versioning information. Both of these aspects will be addressed in future releases.

The ontology's syntax was evaluated through the online RDF validator offered by w3.\footnote{\url{https://www.w3.org/RDF/Validator/}} The tool highlighted no pitfalls in syntax. 

\subsection{Reasoning over the knowledge graph}
The entire knowledge graph was analyzed by the HermiT Reasoner \cite{hermit} to check inconsistencies. The reasoner showed no inconsistencies. The OWL Profile Checker tool (V1.1.0)\footnote{\url{https://github.com/stain/profilechecker}} positions the knowledge graph in the OWL2-DL profile.\footnote{The description logic complexity of the entire knowledge graph is ALCHOIQ (role hierarchies, nominals, inverse properties, qualified cardinality restrictions).}

\subsection{Conversion Evaluation}
\label{sec:conversionevaluation}

To evaluate the quality of the conversion of the algorithms, we manually annotated 112 simulations sampled from some lines of Olderr's dictionary and compared it to the automatic conversion of those same lines using precision, recall and F$_1$ measure of simulacra, reality counterparts, variants, contexts and simulations and specific types of simulations retrieved. 

On average, the conversion achieves a performance of 97\% of precision (micro), recall (micro) and F$_1$ (micro). Table \ref{table:tablemetrics} summarizes the metrics for this evaluation. Errors in the conversion stem from deviations from the dictionary structure, for example when two reality counterparts are provided instead of one.

\begin{figure}[t]
\includegraphics[width=.7\columnwidth]{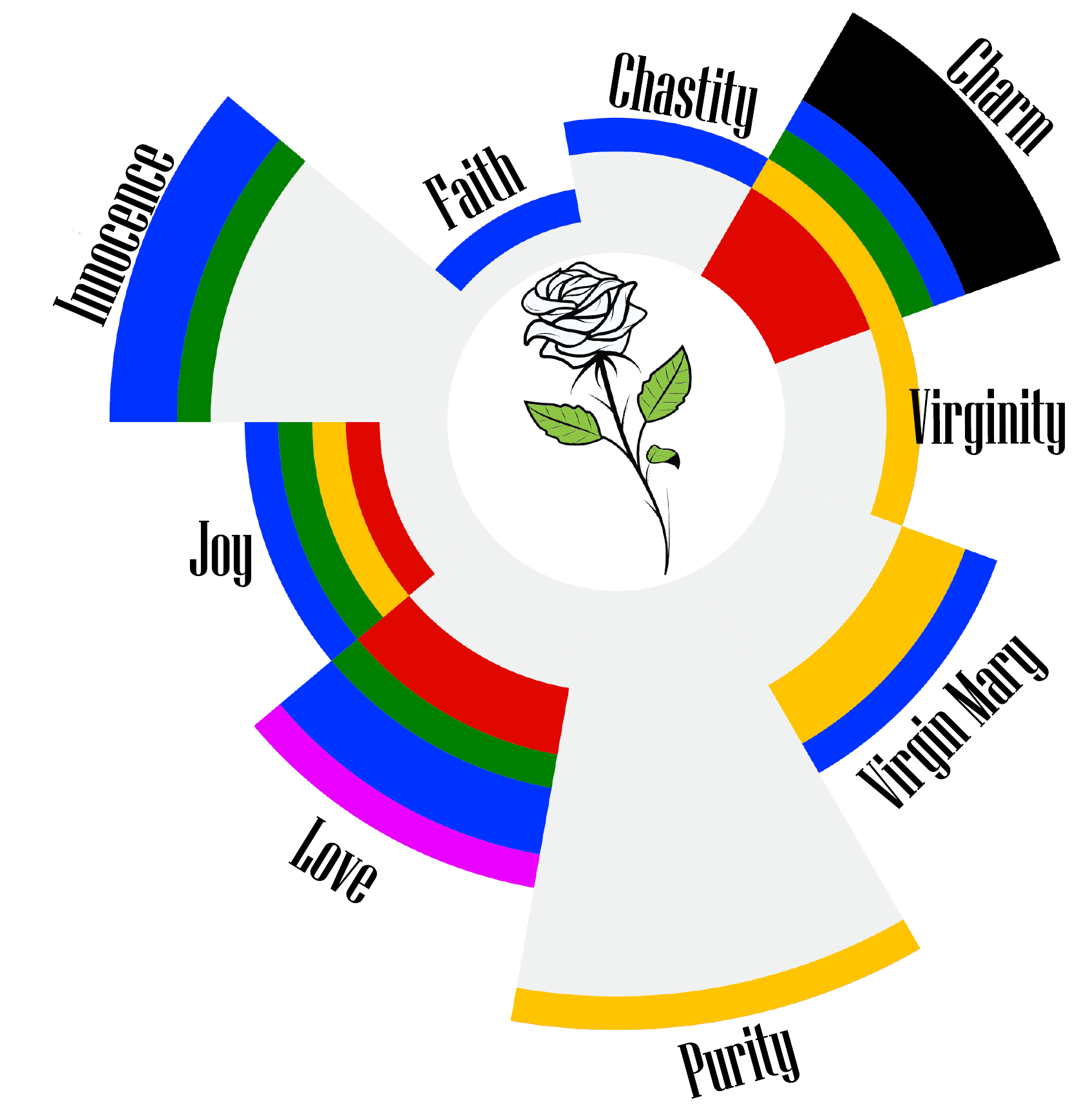}
\caption{Distribution of coloured simulacra across the symbolic meanings of a white rose}
\label{fig:whiterose}
\vspace{-.2cm}
\end{figure}

\section{Case study: white roses}\label{sec:usecase}
In this use case, we show how the knowledge graph can be used to investigate the different meanings of a symbol. We investigate whether there is a correlation between the symbolic meaning of a simulacrum and its colour. We chose white roses as our symbol as they occur across many cultures and in different contexts.  

We extract the symbolic meanings of a white rose from the knowledge graph through the SPARQL query shown in Listing \ref{lst:sparql}. With another series of queries (available on Github), we extract all simulacra that share a symbolic meaning with a white rose that are associated with the colours red, green, black, white, gold, blue, and purple. We created nine sets; one for each symbolic meaning of the white rose, and the coloured extracted simulacra were placed in the set they share the symbolic meaning with (in case of more than one matching of symbolic meaning they were placed in more than one set). For instance, one of the symbolic meanings of a white rose is \textit{purity}. \textit{golden hair}, \textit{white knight}, \textit{white swan} share the same symbolic meaning so they were put in the \textit{purity} set. Only the simulacra's colours were kept, leaving each set with the frequency of colours of simulacra that had the same symbolic meaning.


Figure \ref{fig:whiterose} shows the distribution of coloured simulacra across the different symbolic meanings of a white rose we could extract from our knowledge graph. Every bar represents a symbolic meaning of the white rose. The length of each colour in a bar is proportional to the number of simulacra of that colour that share the same symbolic meaning. As the figure indicates, the white colour is present in almost all symbolic meanings, with a high peak in \textit{purity} and \textit{innocence}. This suggests that the white colour itself brings these symbolic values to white objects. Moreover, apart from the \textit{charm} symbolic meaning, white has the highest percentage of colour in all meanings of white rose except in the meaning for faith which is half white and half blue. 

\begin{lstlisting}[captionpos=b, caption=White rose symbolic meaning SPARQL query, label=lst:sparql,
   basicstyle=\ttfamily,frame=single]
PREFIX kb: <https://w3id.org/simulation/data/> 
PREFIX sim: <https://w3id.org/simulation/ontology/> 

SELECT ?rc WHERE { kb:whiteRose
sim:isSimulacrumOf+/sim:hasRealityCounterpart ?rc .}
\end{lstlisting}

\section{Conclusions and Future Work}\label{sec:conclusions}
In this paper, we present a new ontology and knowledge graph that fill an existing gap in the domain of symbolic knowledge. Our ontology and knowledge graph bring together knowledge from semiotics, mediation, symbolism, and semantic web. We hope this resource will foster research on new quantitative approaches to symbolism that were not possible beforehand.

In future iterations, the knowledge graph could be further enriched by adding information from additional resources describing symbolic information such as Iconclass \cite{Couprie1983IconclassAI}. A SKOS~\cite{skospaper} taxonomy could be generated not only to link simulacra and reality counterparts to their variants, but also in between contexts, to enable more complex queries regarding similar, broader or narrower contexts of simulations. Whilst our ontology and knowledge graph focus on symbolism, metaphorical meanings are partly related to symbolic meaning \cite{coulson2001}, especially when cultural symbols have less conventional, specific meaning. We aim to link our ontology and knowledge graph to
MetaNet\cite{metanet}, an online repository of metaphors and frames. Further integration with external general knowledge graphs such as Wikidata or Framester\cite{framester} would allow interoperability between heterogeneous sources in the context of symbolism, extending the reach of structured symbolic knowledge. This same knowledge could be used to enrich broad knowledge graphs, such as ArCo~\cite{arco}, a knowledge graph of Italian cultural heritage. By automatically assigning potential simulations and symbolic meanings to ArCo objects, scholars and cultural heritage professionals could get access to rich symbolic knowledge, allowing for the discovery of symbolic links between objects that were previously hidden. 

\begin{acks}
This work has been partially funded from the Emilia Romagna Region (grant agreement no. 462 25/03/2019).
 We thank the anonymous reviewers for their constructive feedback.
\end{acks}


\bibliographystyle{ACM-Reference-Format}
\bibliography{KCAPPaper}


\end{document}